%% file: d2s_main_final.tex
\pdfoutput=1

\documentclass[11pt]{article}

\usepackage{naacl2021}

\usepackage{times}
\usepackage{latexsym}
\usepackage{tabularx}
\newcolumntype{Y}{>{\centering\arraybackslash}X}
\usepackage[T1]{fontenc}

\usepackage[utf8]{inputenc}

\usepackage{microtype}
\usepackage{multirow}
\usepackage{enumitem}
\usepackage{graphicx}

\title{D2S: Document-to-Slide Generation Via Query-Based \\ Text Summarization}

\author{Edward Sun\thanks{\hspace{5pt}Work done during internship at IBM Research.} \\
University of Michigan\\
  Michigan, USA \\
  \texttt{edwsun@umich.edu} \\ \And
  Yufang Hou\\
  IBM Research Europe \\
  Dublin, Ireland \\
  \texttt{yhou@ie.ibm.com} \\ \And
  Dakuo Wang\\
  IBM Research \\
  Cambridge, USA \\
  \texttt{dakuo.wang@ibm.com} \\
  \AND
  Yunfeng Zhang \\
  IBM Research \\
  Yorktown, USA \\
  \texttt{zhangyun@us.ibm.com} \\ \And
  Nancy X.R. Wang \\
  IBM Research \\
  Almaden, USA \\
  \texttt{nancywang1991@gmail.com} \\}

\begin{document}
\maketitle
\begin{abstract}

Presentations are critical for communication in all areas of our lives, yet the creation of slide decks is often tedious and time-consuming.
There has been limited research aiming to automate the document-to-slides generation process and all face a critical challenge: no publicly available dataset for training and benchmarking.
In this work, we first contribute a new dataset, SciDuet, consisting of pairs of papers and their corresponding slides decks from recent years’ NLP and ML conferences (e.g., ACL).  
Secondly, we present D2S, a novel system that tackles the document-to-slides task with a two-step approach: 1) Use slide titles to retrieve relevant and engaging text, figures, and tables; 2) Summarize the retrieved context into bullet points  
with long-form question answering. 
Our evaluation suggests that long-form QA outperforms state-of-the-art summarization baselines 
on both automated ROUGE metrics and qualitative human evaluation.

\end{abstract}

\section{Introduction}

\begin{figure}[h]
    \centering
    \includegraphics[width=0.48\textwidth]{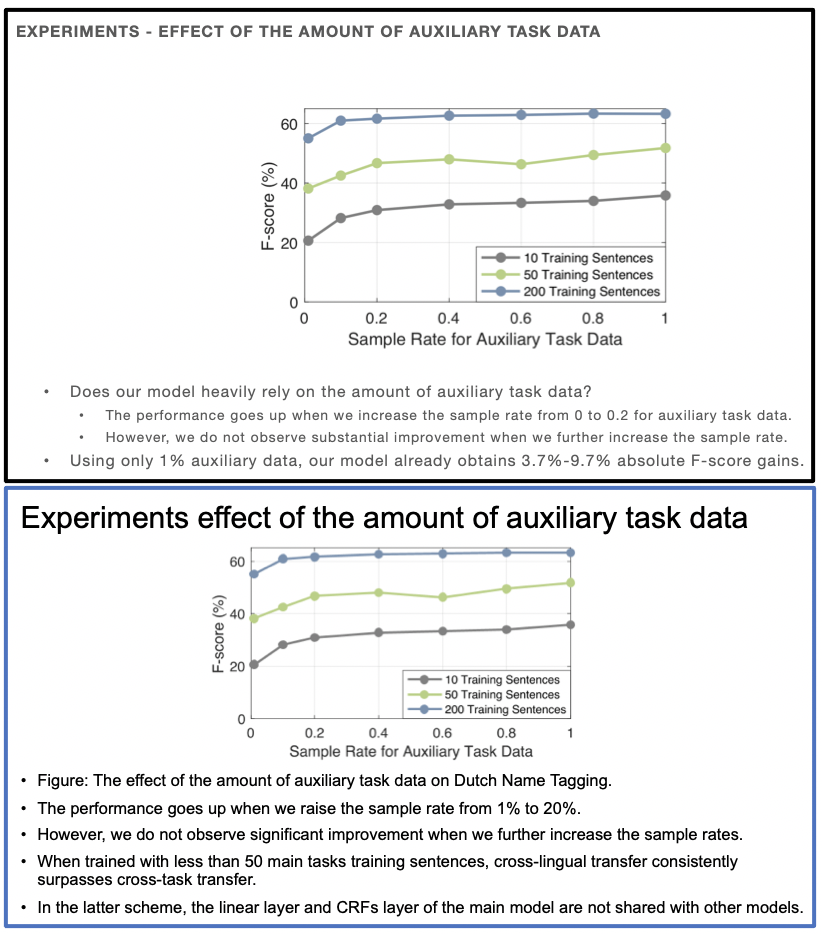}
    \caption{An example slide in SciDuet. TOP is author's original slide; BOTTOM is from our system \textbf{D2S}.}
    \label{fig:sample_slide}
\end{figure}

From business to education to research, presentations are everywhere as they are visually effective in summarizing and explaining bodies of work to the audience~\cite{bartsch2003effectiveness,wang2016people,piorkowski2021ai}.
However, it is tedious and time-consuming to manually create presentation slides~\cite{franco2016automatic}. 

Researchers have proposed various methods to automatically generate presentations from source documents.
For example, \citet{winters2019automatically} suggest heuristic rule-based mechanisms to extract document contents and use those as the generated-slide's content. PPSGen~\cite{hu2014ppsgen} leverages machine learning (ML) approaches to learn a sentence's importance in the document, and extract important sentences as slide's content.

These existing research works have yielded promising progress towards the goal of automated slide generation, but they also face two common limitations: 1) these works primarily rely on extractive-based mechanisms, thus the generated content is merely an aggregation of raw sentences from the document, whereas in real-world slides, the presenter 
frequently uses abstractive summarization; 2) these works assume the presentation slide's title has a one-to-one match to the document's subtitles or section headlines, whereas the presenter in reality often uses new slide titles and creates multiple slides under the same title (e.g., the slides with a \textit{Cont. / Continue} on it).

In our work, we aim to tackle both the limitations.
To achieve this goal, we consider the document-to-slides generation task as a\emph{ Query-Based Single-Document Text Summarization (QSS)} task. 
Our approach leverages recent research developments from \emph{Open-Domain Long-form Question Answering (QA)}.
Specifically, we propose an interactive two-step architecture: in the first step, we allow users to input a short text as the slide title and use a Dense Vector IR module to identify the most relevant sections/sentences as well as figures/tables from the corresponding paper. Then, in the second step, we use a 
QA model to generate the abstractive summary (answer) of the retrieved text based on the given slide title and use this as the final slide text content.

We design a keyword module to extract a hierarchical discourse structure from the paired paper. 
For a given title, we leverage leaf nodes from this tree structure in our IR module to rank paper snippets. 
We further extract related keywords from this structure and integrate them into the QA module.
Experiments demonstrate that the keyword module helps our system to retrieve more relevant context and generate better slide content.

It is worth noting that our system can extract relevant figures and tables for a given title from the source document as well. Figure \ref{fig:sample_slide} (bottom) shows an example of a generated slide from our system.



In addition to our contribution of the novel model architecture, we also contribute a high-quality dataset (\textbf{SciDuet}), which contains 1,088 papers and 10,034 slides. We carefully build this dataset by leveraging a few toolkits for PDF parsing and image/table extraction.  
To the best of our knowledge, this is the first publicly available dataset for the document-to-slides generation task\footnote{Some previous works (SlideSeer~\cite{kan2007slideseer}, PPSGen~\cite{hu2014ppsgen} and ~\cite{wang2017phrase}) described a dataset for training and testing, we could not obtain these datasets with our best ability to search and contact authors.}. 
Our dataset together with the title-based document-to-slide generation task provide a practical testbed for the research field on query-based single-document summarization.
We release the dataset procurement and preprocessing code as well as a portion of \textbf{SciDuet}\footnote{Due to copyright issues, we can only release a portion of our dataset. See Section \ref{dataset} for more details. Other researchers can use our code to construct the full dataset from the original places or extend it with additional data.} at \url{https://github.com/IBM/document2slides}. 


\section{Related Work}
\paragraph{Automated Document-To-Slides Generation}
The early works of automatically generating presentation slides date back to 20 years ago and rely on heuristic rule-based approaches to process information from web searches as slide contents for a user-entered topic~\cite{al2005auto}. 
A recent example, ~\citet{winters2019automatically} used predefined schemas, web sources, and rule-based heuristics to generate random decks based on a single topic.
Among this group of works, different types of rules were used, but they all relied heavily on handcrafted features or heuristics~\cite{shibata2005automatic, prasad2009document, wang2013method}.

More recently, researchers started to leverage machine learning approaches to learn the importance of sentences and key phrases. 
These systems generally consist of a method to rank sentence importance: regression~\cite{hu2014ppsgen, bhandare2016automatic, syamili2017presentation}, random forest~\cite{wang2017phrase}, and deep neural networks~\cite{sefid2019automatic}. And they incorporate another method for sentence selection: integer linear programming~\cite{hu2014ppsgen, sefid2019automatic, bhandare2016automatic, syamili2017presentation} and greedy methods~\cite{wang2017phrase}. 
However, these methods all rely on extractive approaches, which extract raw sentences and phrases from the document as the generated slide content. An abstractive approach based on diverse titles that can summarize document content and generate new phrases and sentences is under-investigated.

\paragraph{Text Summarization}
To support abstractive document-to-slides generation, we refer to and are inspired by the Text Summarization literature. We consider the abstractive document-to-slide generation task as a query-based single-document text summarization (QSS) task. Although there has been increasing interest in constructing large-scale single-document text summarization corpora (CNN/DM \cite{NIPS2015_afdec700}, Newsroom \cite{grusky-etal-2018-newsroom}, XSum \cite{narayan-etal-2018-dont}, TLDR \cite{cachola-etal-2020-tldr}) and developing various approaches to address this task (Pointer Generator \cite{see-etal-2017-get}, Bottom-Up \cite{gehrmann-etal-2018-bottom}, BERTSum \cite{liu2019text}), QSS remains a relatively unexplored field. Most studies on query-based text summarization focus on the multi-document level \cite{Dang2005OverviewOD,10.5555/3016100.3016261} and use extractive approaches \cite{10.1145/3077136.3080690, xu-lapata-2020-coarse}. 
In the scientific literature domain, \newcite{erera-etal-2019-summarization}
apply an unsupervised extractive  approach to generate a summary for each section of a computer science paper. Their system also extracts \emph{task/dataset/metric} entities \cite{hou-etal-2019-identification} which can help users to search relevant papers more efficiently.
In contrast to previous work, we construct a challenging QSS dataset for scientific paper-slide pairs and apply an abstractive approach to generate slide contents for a given slide title. In addition, \newcite{kryscinski-etal-2019-neural} argues that future research on summarization should shift from ``general-purpose summarization'' to constrained settings. The new dataset and task we proposed provide the practical testbed to this end.

\paragraph{Open-Domain Long-Form Question Answering}
Our work is motivated by 
the recent advancements in open-domain long-form question answering task, in which the answers are long and can span multiple sentences (ELI5 \cite{fan-etal-2019-eli5}, NQ \cite{kwiatkowski-etal-2019-natural}). Specifically, we consider the user-centered slide titles as questions and the paper document as the corpus. 
We 
use information retrieval (IR) to collect the most relevant text snippets from the paper for a given title before passing this to a QA module for sequence-to-sequence generation.
We 
further improve the QA module by integrating title-specific key phrases to guide the model to generate slide content. In comparison to ELI5 and NQ, the questions in the slide generation task are shorter; and a significant proportion of the reference answers (slide contents) contain tables and figures directly from the paper, which then requires particular consideration. 


\section{SciDuet Dataset Construction}
\label{dataset}
\paragraph{Data Sources}
The SciDuet (SCIentific DocUment slidE maTch) dataset comprises of paper-slide pairs scraped from online anthologies of International Conference on Machine Learning (ICML'19), Neural Information Processing Systems (NeurIPS'18\&'19),  
and Association for Computational Linguistics (since ACL'79) conferences. 
We focus only on machine learning conferences as their papers have highly specialized vocabulary; we want to test the limits of language generation models on this challenging task. Nevertheless, these generic procuration methods (web-scraping) can be applied to other domains with structured archives. 

\paragraph{Data Processing}
Text on papers was extracted through Grobid \cite{GROBID}. Figures and captions were extracted through pdffigures2 \cite{clark2016pdffigures}. Text on slides was extracted through IBM Watson Discovery package\footnote{https://www.ibm.com/cloud/watson-discovery} and OCR by \emph{pytesseract}.\footnote{https://pypi.org/project/pytesseract} 
Figures and tables that appear on slides and papers were linked through multiscale template matching by OpenCV. Further dataset cleaning was performed with standard string-based heuristics on sentence building, equation and floating caption removal, and duplicate line deletion.  


\paragraph{Dataset Statistics and Analysis}
SciDuet has 952--55--81 paper-slide pairs in the Train--Dev--Test split. 
We publicly release SciDuet-ACL which is constructed from ACL Anthology. It contains the full Dev and Test sets, and a portion of the Train dataset. Note that although we cannot release the whole training dataset due to copyright issues, researchers can still use our released data procurement code to generate the training dataset from the online ICML/NeurIPS anthologies.

Table \ref{tab:datastat1} shows the statistics of the dataset after excluding figures and tables from slide contents. In the training dataset, 70\% of slide titles have fewer than five tokens, and 59\% of slide contents have fewer than 50 tokens.\footnote{Note that in Table \ref{tab:datastat1}, the Dev set has a slightly longer \emph{SC-len} than the Train/Test sets. This is because there are two papers in the Dev set whose slides contain a lot of words. \emph{ST-Len} in the Dev set is 56 after removing these two papers.}

\begin{table}[t]
    \centering
    \begin{tabular}{|c|c|c|c|c|}
    \hline
         & \#papers & \#slides &ST-len&SC-len\\ \hline
      train   & 952& 8,123&3.6&55.1\\ \hline
      dev &55&733&3.16&63.4 \\ \hline
      test &81&1,178&3.4& 52.3\\ \hline
    \end{tabular}
    \caption{Dataset statistic. `ST-len` and `SC-len ` indicate the average token length for slide titles and slide contents, respectively.}
    \label{tab:datastat1}
\end{table}


\begin{table}[t]
    \centering
    \begin{tabular}{|l|c|c|c|} 
    \hline
    &\multicolumn{3}{c|}{\% of novel n-grams} \\ \hline 
         & unigrams & bigrams &trigrams\\ \hline 
      SlideTitle   & 35.1& 71.9&87.4\\ \hline
      SlideContent &22.9&64.1&84.7 \\ \hline
    \end{tabular}
    \caption{The average proportion of novel n-grams for slide titles and slide contents in the training dataset.}
    \label{tab:datastat2}
\end{table}

\begin{figure*}[t]
    \centering
    \includegraphics[width=1.0\textwidth]{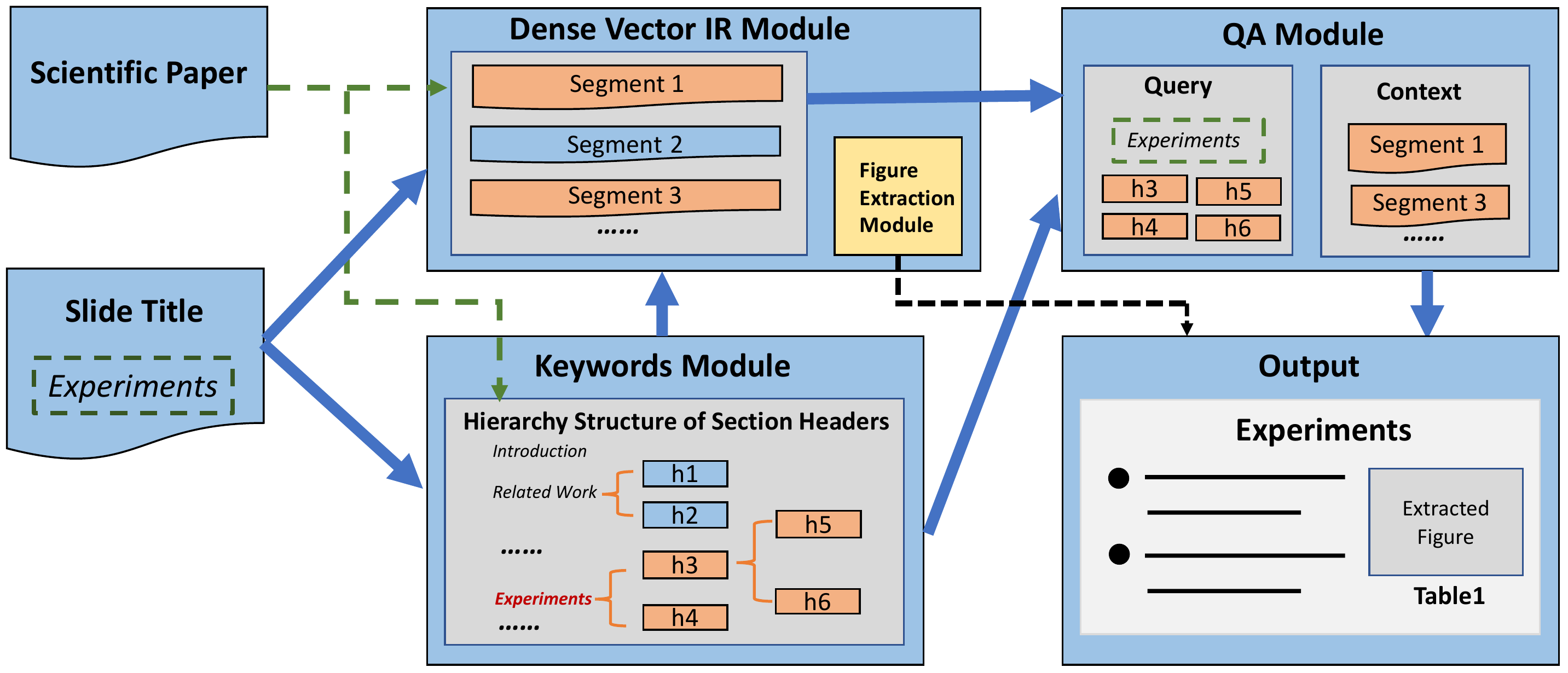}
    \caption{System architecture of our D2S framework.}
    \label{fig:datastat3}
\end{figure*}

We also calculate the novel n-grams for slide titles and slide contents compared to the corresponding papers in the training dataset (Table~\ref{tab:datastat2}). It seems that slide titles contain a higher proportion of novel n-grams compared to slide contents. Some examples of novel n-grams in slide titles are: \emph{recap}, \emph{motivation}, \emph{future directions}, \emph{key question}, \emph{main idea}, and \emph{final remarks}.
Additionally, we found that only 11\% of slide titles can match to the section and subsection headings in the corresponding papers.


\section{D2S Framework}
\label{sec:framework}
We consider document-to-slide generation as a closed-domain long-form question answering problem. Closed domain means the supporting context is limited to the paired paper. While traditional open-domain QA has specific questions, nearly 40\% of our slide titles are generic (e.g., \emph{take home message}, \emph{results}). To generate meaningful slide contents (answers) for these titles (generic questions), we use title-like keywords to guide the system to retrieve and generate key bullet points for both generic titles and the specific keywords.

The system framework is illustrated in Figure~\ref{fig:datastat3}. 
Below, we describe each module in detail. 

\subsection{Keyword Module}
The inspiration for our Keyword Module is that paper often has a hierarchy structure and unspecified weak
titles (e.g., \emph{Experiments} or \emph{Results}). 
We define \textit{weak titles} as undescriptive generic titles nearly identical to section headers. 
The problem with these generic section headers is the length of their sections. Human presenters know to write content that spans the entire section. E.g., one may make brief comments on each subsection for a long \emph{Experiments} section.
For that, we use the keyword module to construct a parent-child tree of section titles and subsection headings. We use this hierarchical discourse structure to aid our D2S system to improve information retrieval (Section \ref{sec:ir}) and slide content generation (Section \ref{sec:qa}).



\subsection{Dense IR Module}
\label{sec:ir}
Recent research has proposed various embedding-based retrieval  approaches \cite{Guu2020REALMRL, lee2019latent, karpukhin-etal-2020-dense} which outperform traditional IR methods like BM25. In our work, we integrate 
the leaf nodes of the parent-child trees from the keyword module into the reranking function of a dense vector IR system based on a distilled BERT miniature \cite{Turc2019WellReadSL}.


Without gold passage annotations, we train a dense vector IR model to minimize the cross-entropy loss of titles to their original content (taken from the original slides) 
because of 
their similarity to paper snippets.
For a given title $t$, we randomly choose slide contents from other slides with different titles as the negative samples.

We precompute
vector representations 
for all paper snippets (4 sentence passages) with 
the pre-trained IR model. 
We then apply this model to
compute a same-dimension dense vector representation for slide titles. Pairwise inner products are computed between the vectors of all snippets from a paper and the vector of a slide title. We use these inner products to measure the similarity between all title-snippet pairs, and we rank the paper passage candidates in terms of relevance to a given title with the help of Maximum Inner Product Search ~\cite{johnson2019billion}. The top ten candidates are selected as input's context to the QA Module.
We further improve the IR re-ranking with 
extracted section titles and subsection headings (keywords) from the Keyword Module.
We design a weighted ranking function with vector representations of titles, passage texts, and the leaf node keywords: 
\[\alpha(\mathsf{emb}_{title}\cdot \mathsf{emb}_{text})+(1-\alpha)(\mathsf{emb}_{title}\cdot \mathsf{emb}_{text_{kw}})\]
where $\mathsf{emb}_{title}$, $\mathsf{emb}_{text}$, and $\mathsf{emb}_{text_{kw}}$ are the embedding vectors based on the pre-trained IR model for a given title, a text snippet, and the leaf node keyword from the keyword module which contains the text snippet, respectively. We find from the dev set that $\alpha=0.75$ is optimal. 
Experiments in Section \ref{sec:ir_result} shows that
this ranking function can help our system become more header-aware and robust. 



\subsection{QA Module}
\label{sec:qa}
The QA module in our D2S system combines slide title and the corresponding keywords as the query.  It takes the concatenation of the top ten ranked text snippets from the IR module as the context. 

We match a title to a set of keywords using the parent-child hierarchy from the keyword module. 
Note that this hierarchy is not limited to core sections $(1,2,3,\dots)$, but can also be leveraged for all paper header matches x.x.x Specifically, if a title $t$ matches 
with a header 2.1 (Levenshtein ratio $\geq 0.9$), then we will include header 2.1 as well as all of its recursive children (e.g., 2.1.x, 2.1.x.x) as keywords for the QA module. 
It is worth noting that not every title has corresponding keywords. 

Our QA model is a fine-tuned BART \cite{lewis-etal-2020-bart}. We encode the query and the context in the format of 
 ``\{\emph{title}[SEP1]{keywords}[SEP2]context\}''.
Keywords were embedded sequentially as a comma-separated list into the input following the slide title. We hypothesize that integrating keywords into the query can help our model pay attention towards relevant important context across all retrieved text fragments when generating slide content. This is indeed effective when the slide titles are aligned with broader sections, such as “Results”. In practice, embedding keywords helps the model in not just summarizing the top-ranked paragraphs, but also paying attention to additional paragraphs relevant to the broad topic.

We fine-tune our QA model using filtered training data. Filtering is done because the process of humans generating slides from a paper is highly creative and subjective to each author's unique style. Some may include anecdotes or details outside the paired paper. These underivable lines, if not filtered, may hinder the QA module's performance on generating faithful sentences from the paired paper. 
Our experiments support this speculation.\footnote{Note that we always evaluate our model's performance on the unfiltered dataset (Section~\ref{sec:data-filter-experiment}).} 

\paragraph{Training Data Filtering}
Due to the abstractive nature of slides, it is difficult to filter out slide content that is underivable from the paper content.
No existing automated metrics can be used as a threshold to differentiate the derivable or underivable lines.
To approach this, we performed manual gold standard annotations on 200 lines from slides to determine derivability.  This led to the development of a Random Forest Classifier trained on the majority voting decision of annotators for 50 lines and tested on the remaining 150 lines. The classifier feature space is a combination of ROUGE-(1, 2, L) recall, precision, and F-scores. We apply this classifier to the original training set to filter out slide content that likely cannot be derived from the paired papers.\footnote{The derivability annotations together with the trained classifier can be accessed on our GitHub.}


\subsection{Figure Extraction Module}
Slide decks are incomplete without good visual graphics to keep the audience attentive and engaged. Our D2S system adaptively selects connected figures and tables from the paper to build a holistic slide generation process.
Our implementation is simple, yet effective. It reuses the dense vector IR module (Section \ref{sec:ir}) to compute vector similarities between the captions of figures/tables and the slide title (with the extended keywords if applicable). Figures and tables are then ranked and a final recommendation set is formed and presented to the user. This simulates an interactive figure recommendation system embedded in D2s.

\section{Experimental Setup}

\paragraph{Implementation Details}
All training was done on two 16GB P100 GPUs in parallel on PyTorch. Our code adapts the transformer models from HuggingFace~\cite{wolf2020transformers}. All hyperparameters are fine-tuned on the dev set. 
A distilled uncased Bert miniature with 8-layers, 768 hidden units, and 12 attention heads was trained and used to perform IR. The BERT model computes all sentence embeddings in 128-dimensional vectors.
Our QA model was fine-tuned over \emph{BART-Large-CNN}, a BART model pre-trained on the CNN-Dailymail dataset. Pilot experiments showed that \emph{BART-Large-CNN} outperforms \emph{BART-Large} and other state-of-the-art pre-trained language generation models on the dev dataset. The BART model used the AdamW optimizer and a linear decreasing learning rate scheduler.

During testing, we apply our trained QA model with the lowest dev loss to the testing dataset. Note that we do not apply any content filtering on the testing data. The QA models generate the predicted slide content using beam search with beam 8 and no repeated trigrams. We use the dev dataset to tune the minimum and maximum token lengths for the output of the QA model. 

\paragraph{Evaluation}
We evaluate our IR model using IDF-recall, which computes the proportion of words in the original slide text in the retrieved context weighted by their inverse document frequency. This metric gives more focuses to important words. For adaptive figure selection, we report the top-(1, 3, 5) precision. 
Finally, for slide text content generation, we use ROUGE as the automatic evaluation metric \cite{lin-2004-rouge}. We also carried out human evaluation to assess the D2S system's performance on slide text generation.

\section{Evaluation on IR and Figure Selection}
\label{sec:ir_result}
\paragraph{Results on Dense IR}
For a given slide title, the goal of IR is to identify relevant information from the paired paper for the downstream generation model.
We compare our IR model (\emph{Dense-Mix IR}) described in Section \ref{sec:ir} to a few baselines. \emph{Classical IR (BM25)} is based on sparse word matching which uses the BM25 similarity function. \emph{Dense-Text IR} and \emph{Dense-Keyword IR} are variants of our \emph{Dense-Mix IR} model with different ranking functions ($\alpha$ equals 1 for \emph{Dense-Text IR} and 0 for \emph{Dense-Keyword IR}). 

All experiments are evaluated on the test set.
The IDF-recall scores for each IR method are as follows: \emph{Classical IR (BM25)} = 0.5112, \emph{Dense-Text IR} = 0.5476, \emph{Dense-Keyword IR} = 0.5175, and \emph{Dense-Mix IR} = 0.5556.

The experiments indicate the dense IR model outperforming the classical IR approach and an $\alpha=0.75$-weighted mix dense IR model outperforming other dense IR models that rank exclusively by text or keywords.

These results support the design decision of using embedding-based IR and re-ranking based on both text snippets and keywords. 
We attribute the success of the \emph{Dense-Mix IR} model to increased section header awareness. Header-awareness leads to better retrieval in cases where the title corresponds well with section headers. The drawback of ranking solely on keywords is in the case when the dense IR module cannot differentiate between passages with the same header. This leads us to find the right balance ($\alpha=0.75$) between \emph{Dense-Text IR} and \emph{Dense-Keyword IR}.

\paragraph{Results on Figure Selection}
We evaluate figure selection based on the set of the testing slides which contain figures/tables from the paired papers. 
The results of the adaptive figure selection are promising. 
It achieves 0.38, 0.60, and 0.77 on $p@1$, $p@3$, and $p@5$, respectively.
This suggests our system is holistic and capable of displaying figures and tables for slides that the original author chose.



\section{Evaluation on Slide Text Generation}
\subsection{Baselines and BARTKeyword}
Below we describe the technical details of the two baselines as well as our QA module (\emph{BARTKeyword}) for slide text generation.

\paragraph{BertSummExt} From~\cite{liu2019text}, the model is fine-tuned to the retrieved context on our unfiltered training dataset.
For a given title and the retrieved context based on our IR model, the model
extracts important sentences from the context as the slide text content.
Note that performance was lowered with filtering, which differs from other models. We suspect that the extractive model depends on output text lengths. 
Filtering reduces the ground truth token length, which in turn, makes the generated output also shorter, leading to a marginally higher precision at greater cost in recall.
Hyperparameters are reused from ~\cite{liu2019text} and training continues from the best pre-trained weights of the CNN/Daily-Mail task. This maintains consistency with the Bart models, which were also pre-trained on CNN/DM.
\paragraph{BARTSumm} A BART summarization model fine-tuned on the filtered dataset. We use a batch size of 4 with an initial learning rate of $5e$-$5$. We set the maximum input token length at 1024, which is approximately the same length as the retrieved context (10 paper snippets $\approx$ 40 sentences, each sentence $\approx$ 25 tokens). Min and max output token lengths were found to be 50 and 128.


\paragraph{Our Method (BARTKeyword)} 
This is our proposed slide generation model as described in Section \ref{sec:qa}.
We fine-tune our QA model on the filtered dataset with a batch size of 4 and an initial learning rate of $5e$-$5$. The maximum input token length was also set to 1024. Dev set tuned min and max token lengths were found to be 64 and 128.
\subsection{Results and Discussion}

\subsubsection{Automated Evaluation}

\begin{table*}
\centering
\begin{tabular}{|l|c|c|c|c|c|c|c|c|c|}
\hline
\multirow{2}{*}{\textbf{Summarization Model}} &  \multicolumn{3}{c|}{\textbf{ROUGE-1}} &  \multicolumn{3}{c|}{\textbf{ROUGE-2}}&  \multicolumn{3}{c|}{\textbf{ROUGE-L}}\\ \cline{2-10}
&P&R&F&P&R&F&P&R&F\\
\hline 
 \multicolumn{10}{|c|} {\emph{Classical IR (BM25)}}\\ \hline
BertSummExt & 14.26 & 24.07 & 15.89 & 2.59 & 4.46 & 2.86 & 12.89 & 21.70 & 14.31 \\ \hline
BARTSumm & 15.75&	23.40&	16.92&	2.94&	4.12&	3.11&	14.18&	20.99	&15.55 \\\hline
BARTKeyword (ours) & 17.15 &27.98 & 19.06 & 4.08 & 6.52 & 4.52 & 16.29 & 24.88 & 18.12 \\\hline

 \multicolumn{10}{|c|} {\emph{Dense-Mix IR (ours)}}\\ \hline
BertSummExt & 15.47 & 25.74 & 17.16 & 3.14 & 5.24 & 3.47 & 13.97 & 23.29 & 15.48 \\ \hline
BARTSumm & 16.62&	26.10&	18.15&	3.35&	5.16&	3.63&	15.00&	23.28	&16.73 \\\hline
BARTKeyword (ours) & \textbf{18.30} & \textbf{30.31} & \textbf{20.47} & \textbf{4.73} & \textbf{7.79} & \textbf{5.26} & \textbf{16.86} & \textbf{27.21} & \textbf{19.08} \\\hline

\end{tabular}

\caption{ROUGE scores of our \emph{BARTKeyword} QA model compared to other summarization baselines based on different IR approaches.}
\label{tab:full-rouge}
\end{table*}

We use ROUGE scores to evaluate the generated content with regard to the ground-truth slide content.
Overall,  our \emph{Dense-Mix IR} approach provides better context for the downstream summarization models.
In general, our BARTKeyword model is superior to the abstractive and extractive summarization models in all ROUGE metrics (1/2/L) based on different IR approaches as shown in Table~\ref{tab:full-rouge}.
Additionally, the abstractive summarization model performs better than the extractive model.
Altogether, this shows the importance of adopting an abstractive approach as well as incorporating the slide title and keywords as additional context for better slide generation from scientific papers.

\paragraph{Human-Generated Slides (Non-Author)}
\label{sec:human-generated-slides}

\begin{table}
\centering
\begin{tabularx}{\columnwidth}{|l|Y|Y|Y|}
\hline
\textbf{Generator} & \textbf{R-1} & \textbf{R-2} & \textbf{R-L}\\
\hline 
Humans & 26.41& \textbf{8.66} & 24.68 \\\hline
D2S  & \textbf{27.75} & 8.30 & \textbf{24.69} \\\hline

\end{tabularx}
\caption{ROUGE F-scores for non-author generated slides in comparison to our \emph{D2S} system.}

\label{tab:human-baseline}
\end{table}

As presentation generation is a highly subjective task, we wanted to estimate the expected ROUGE score that a non-author (but subject domain expert) human may be able to obtain by generating slides from the paper. 
In total, three authors of this paper each randomly selected and annotated one paper (either from dev or test), and another common paper (\textit{Paper ID: 960}), thus in total four papers have non-author human-generated slides.\footnote{Manually generated slides available on our GitHub.} The procedure we followed was: read the paper thoroughly, and then for each slide's original title, generate high-quality slide content using the content from the paper. 
The high quality of our non-author experts generated slides can be  demonstrated through the high scores given for the human-generated slides in the human evaluation (Section \ref{sec:human-evaluation}). 


Table~\ref{tab:human-baseline} shows the results of ROUGE F-score for non-author generated slides compared to our D2S system. It is interesting to see that our model's performance is similar to or sometimes better than the non-author generated ones. 
The task of generating slides from a given paper is indeed difficult even for subject domain experts, which is quite a common task in ``research paper reading groups''. It is easy for humans to miss important phrases and nuances, which may have resulted in the lower score compared to the model.

In general, the low human annotator ROUGE F-score shown in Table~\ref{tab:human-baseline} reflects the difficulty and subjectivity of the task.
This result also provides a reasonable performance ceiling for our dataset.

\subsubsection{Human Evaluation}
\label{sec:human-evaluation}
\paragraph{Four Models}  As suggested in the ACL'20 Best Paper ~\cite{ribeiro2020beyond}, automatic evaluation metrics alone cannot accurately estimate the performance of an NLP model. 
In addition to the automated evaluation, we also conducted a human evaluation to ask raters to evaluate the slides generated by \textbf{BARTKeyword} (our model), by baseline models (both \textbf{BARTSumm} and \textbf{BertSummExt}) based on \emph{Dense-Mix IR},
and by the non-author human experts (\textbf{Human}). 


\paragraph{Participant} 
The human evaluation task involves reading and rating 
slides from the ACL Anthology.
We noted that some technical background was required, so we recruited machine learning researchers and students $(N=23)$ with snowball sampling.
These participants come from several IT companies and universities. Among them: 10 have more than 3 years of ML experience; 7 have more than 1 year; 13 actively work on NLP projects; and 7 know the basic concepts of NLP.

\paragraph{Dataset} In the human evaluation, 
we use 81 papers from the test set.
We filter out papers with fewer than 8 slides, as each rater will do 8 rounds in an experiment, leaving 71 papers in the set.

\paragraph{Task} We follow prior works' practices of recruiting human raters to evaluate model-generated contents~\cite{wang2021cass}. 
For each rater, we randomly select two papers, one from the former four papers, and another one from the test set. 
For each paper, we again randomly select four slides, thus each participant complete eight rounds of evaluation (2 papers $\times$ 4 slides).
In each round, a participant rates one slide's various versions from different approaches with reference to the original author's slide and paper. The participants rate along three dimensions with a 6-point Likert scale (1 strongly disagree to 6 strongly agree):

\begin{itemize}[noitemsep]
    \item \textbf{Readability}: The generated slide content is coherent, concise, and grammatically correct;
    \item \textbf{Informativeness}: The generated slide provides sufficient and necessary information that corresponds to the given slide title, regardless of its similarity to the original slide;
    \item \textbf{Consistency}: The generated slide content is similar to the original author's reference slide.
\end{itemize}


\paragraph{Result} Ratings on the same model's slides are aggregated into an average, resulting in three scores for each of the four models (three systems plus \textbf{Human}). ANOVA tests are used for each dimension (Greenhouse-Geisser correction applied when needed) to compare the models' performances, and a post hoc pairwise comparisons with Tukey's honest significance difference (HSD) test~\cite{field2009discovering}.

\begin{figure}
    \centering
    \includegraphics{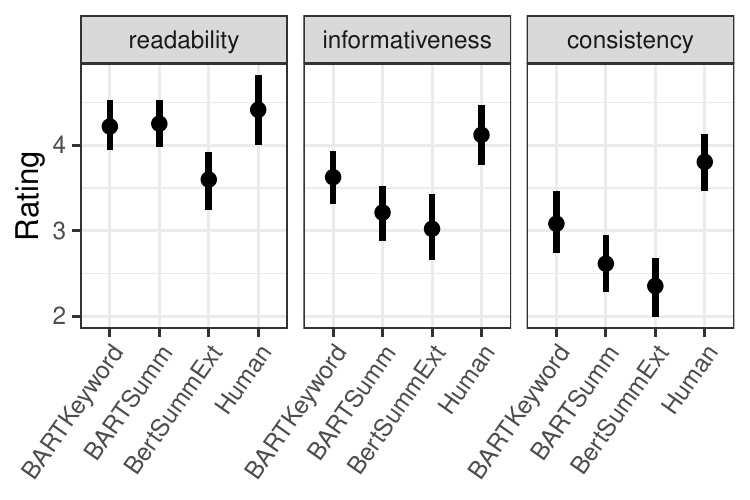}
    \caption{Average rating given by participants to each method across three dimensions. Error bars represent 95\% confidence intervals.}
    \label{fig:ratings}
\end{figure}

Results show that for the \textbf{Readability} dimension (Figure~\ref{fig:ratings}), the BertSummExt model performs significantly worse than the other three models ($F(1.77, 39.04)=6.80$, $p=.004$), and that between the three models there is no significant difference. This result suggests that even though the extraction-based methods use grammatically correct sentences from the original paper, the human raters do not think the content is coherent or concise; however, it also indicates that summarization-based models can achieve fairly high readability.

The most \textbf{Informative} slides were generated by humans ($F(1.59, 35.09)=13.10$, $p<.001$).
But BARTKeyword (our model) came in second and outperformed BertSummExt significantly  ($t(66)=3.171$, $p=.012$) and BARTSumm insignificantly ($t(66)=2.171$, $p=.142$). Our model is also the only ML model rated above 3.5 (the midpoint of the 6-point Likert scale), meaning on average, participants agree that the model is informative.

Regarding the \textbf{Consistency} between the generated slide content and the author's original slide, there is a significant difference in ratings across methods, $F(1.68, 37.03)=30.30$, $p<.001$. Human-generated slides outperformed the ML models again in this metric, but BARTKeyword also significantly outperformed the other two: $t(66)=4.453$, $p<.001$ vs BertSummExt, and $t(66)=2.858$, $p=0.028$ vs BARTSumm. This indicates that our model provides a SOTA performance in the consistency dimension, but there is space to improve to reach the human level.


\section{System Analysis}
In this section, we carry out additional experiments to better understand the effectiveness of different components in our system. 
\paragraph{IR-Oracle} To estimate an upper-bound of the ROUGE score, we design an IR model to locate the best context possible for retrieval. For each line in the ground-truth slide content, we retrieve the most related sentence from the entire paper scored by a weighted ROUGE score. This oracle model sees information that would not be available in a regular slide generation system, which only has the title as input.
Similar to what was shown in the human annotation experiment (Table~\ref{tab:human-baseline}), the F-score in all ROUGE metrics remain below 40 (Table~\ref{tab:addl-expts}, row \emph{Oracle-IR}), demonstrating the subjectiveness of the task and providing context for the level of performance achieved by the D2S System.

\paragraph{Effect of keywords in Summarization}
Section \ref{sec:ir_result} shows that our keywords aware Dense-Mix IR model achieves the best IDF-recall score on the test dataset. Here we test the effect of keywords in the QA module. Table \ref{tab:addl-expts} shows that removing keywords from BARTKeyword (\emph{BaseQA}) leads to performance degradation. It seems that the extracted keywords for a given title can help our model to locate relevant context from all retrieved text snippets and generate better content.



\label{sec:data-filter-experiment}
\paragraph{Effect of Dataset Filtering in Summarization} 
We also test the effect of filtering the training dataset in the QA module. Table \ref{tab:addl-expts} shows that training \emph{BARTKeyword} on the filtered training dataset (described in Section \ref{sec:qa}) helps improve performance in the unfiltered test set. This is likely due to the reduction of noisy text that cannot be generated from the document, allowing the model to learn to synthesize information from the text without trying to hallucinate new information. 


\begin{table}
\centering
\begin{tabularx}{\columnwidth}{|l|Y|Y|Y|}
\hline
&\textbf{R-1}&\textbf{R-2}&\textbf{R-L}\\
\hline
 \multicolumn{4}{|c|} {\emph{BARTKeyword}}\\ \hline
Oracle-IR & 36.32 & 16.99 & 36.59 \\\hline 
Dense-Mix IR & 20.47 & 5.26 & 19.08 \\ \hline
 \multicolumn{4}{|c|} {\emph{Dense-Mix IR}}\\ \hline
BARTKeyword & 20.47 & 5.26 & 19.08 \\\hline
BaseQA &  20.19 & 5.04 & 18.81 \\\hline
\multicolumn{4}{|c|} {\emph{Dense-Mix IR + BARTKeyword}}\\ \hline
Filter &  20.47 & 5.26 & 19.08 \\\hline
Unfilter & 20.02 & 4.94 & 18.64  \\\hline

\end{tabularx}
\caption{ROUGE F-scores of system variations.}
\label{tab:addl-expts}
\end{table}

\section{Error Analysis}
To gain additional insights into our model’s performance, we carried out a qualitative error analysis to check the common errors in our best system \emph{(Dense-Mix IR + BARTKeyword)}. We sampled 20 slides that received lower rating scores (rating score $<$ 3 in at least one dimension) in our human evaluation experiment (Section \ref{sec:human-evaluation}). One author of this paper carefully checked each generated slide content and compared it to the original paper/slide. 

In general, we found that most errors are due to off-topic content. For instance, given a slide title ``\emph{Future Work}'', our model might generate sentences that summarize the major
contributions of
the corresponding paper but do not discuss next steps. We also observed that occasionally our model hallucinates content which is not supported by the corresponding paper. Normally, this happens after the model selects an example sentence from the paper and the sentence's content is very different from its surrounding context. For instance, a paper uses an example sentence ``\emph{Which cinemas screen Star Wars tonight?}'' to illustrate a new approach to capture intents/slots in conversations. Then for the slide title ``\emph{Reminder Q\&A Data}'', our model generates ``\emph{Which cinemas screen Star Wars tonight? Which movie theater plays Star Wars at 8 p.m. on December 18?}''. Here, the second sentence is 
a hallucination error. 

We use the novel n-grams to measure the ``abstractiveness'' of the generated slide contents. On the testing dataset, we found that the original slide contents contain a much higher proportion of novel n-grams compared to the automatically generated ones (e.g., 24.2\% vs. 3.1\% for novel unigrams, and 66.5\% vs. 14.2\% for novel bigrams). This indicates that the generated slide contents from our model are still mostly ``extractive''.

\section{Conclusion}
This project aims to automatically generate presentation slides from paper documents. 
The problem is framed as a query-based
single-document summarization task. Inspired by recent work on open-domain long-form QA, we design a keyword-aware framework (D2S) to tackle this challenge.  
Both automated and human evaluations suggest that our system outperforms a few strong baselines and can be served as a benchmark for the document-to-slide challenge. We release the dataset (SciDuet) and code in hopes it can foster future work.


 

\section*{Acknowledgments}
The authors appreciate the valuable feedback from the anonymous reviewers.
We also thank our friends and colleagues who participated in our human evaluation study.

\bibliography{anthology,acl2020}
\bibliographystyle{acl_natbib}

 \newpage
 \newpage

\appendix
\section{Appendices}
\label{sec:appendix}
\input{d2s_appendix}

\end{document}

%% file: d2s_appendix.tex
\begin{table*}[th]
\centering
\begin{tabular}{|l|l||c|c|c|}
\hline
\textbf{Paper(s)} & \textbf{Generator} & \textbf{ROUGE-1} & \textbf{ROUGE-2} & \textbf{ROUGE-L}\\
\hline \hline
960 & Humans & 23.91 (2.97) & 6.55 (0.79) & 24.23 (2.03) \\ \hline
960 & Human-best & 28.10 & 7.66 & \textbf{27.10} \\\hline
960 & BARTKeyword (ours) & \textbf{29.48} & \textbf{8.16} & 26.12 \\\hline \hline
All & Humans & 26.41 (4.80) & \textbf{8.66 (2.24)} & 24.68 (2.03) \\\hline
All & BARTKeyword (ours) & \textbf{27.75 (1.62)} & 8.30 (0.36) & \textbf{24.69 (1.18)} \\\hline
\end{tabular}
\caption{ROUGE  F-scores  for  non-author  generated slides for four papers in comparison to our D2S system. }
\label{tab:human-four}
\end{table*}

\begin{figure*}[ht]
    \centering
    \includegraphics[width=0.74\textwidth]{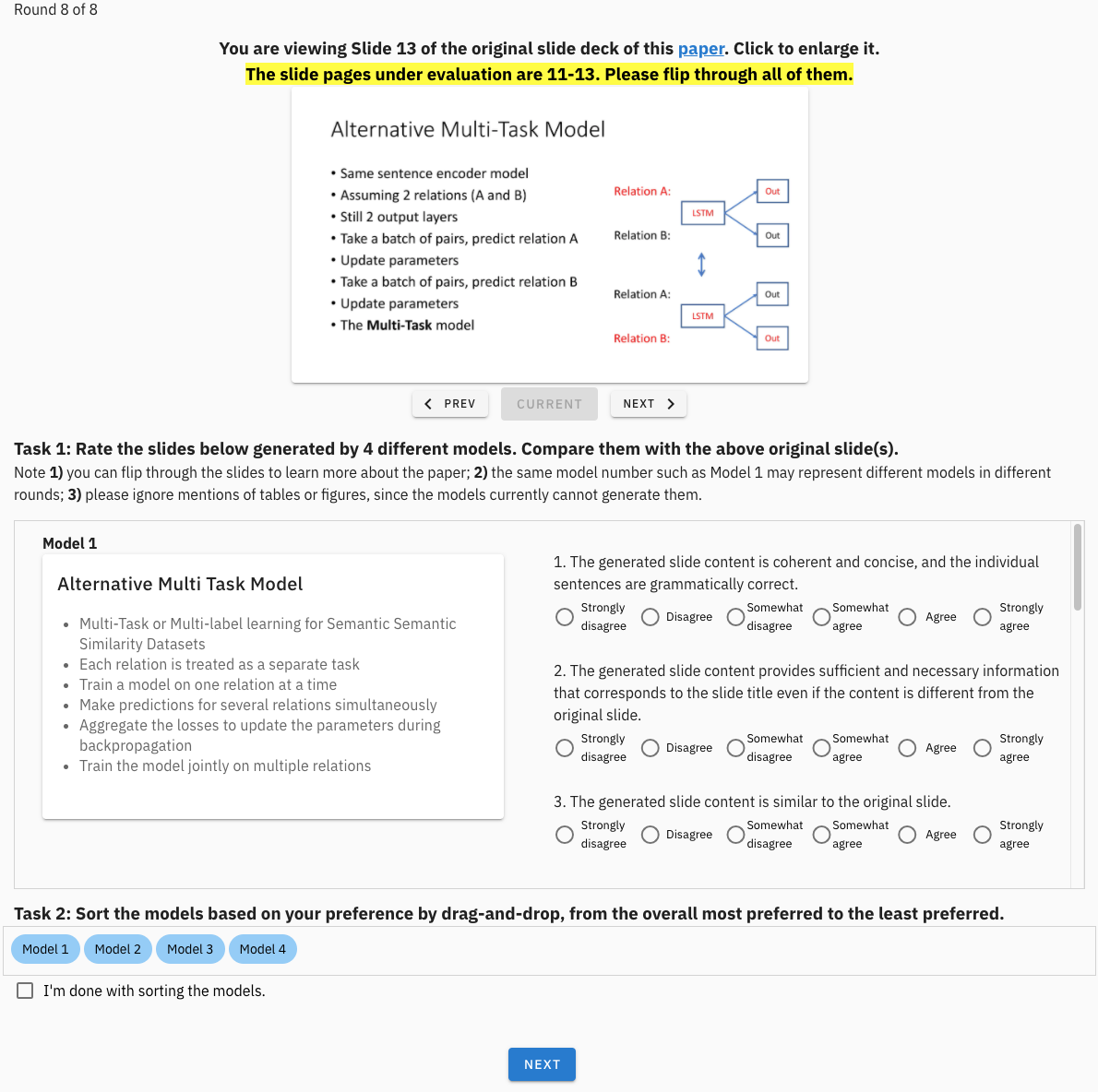}
    \caption{A screenshot of the web survey.}
    \label{fig:survey_screenshot}
\end{figure*}

\subsection{Model Training and Parameters}
The BART-based models converge between 0-10 epochs, depending on the learning rate, and take around 5-10 hours. All learning rates are selected from the following set: $(1e$-$5,2e$-$5,5e$-$5,1e$-$4,2e$-$4,5e$-$4)$. Min-max token output lengths were tuned from the following set: $(32$-$128,32$-$256, 50$-$128,50$-$256,64$-$128,64$-$200,64$-$256)$. Batch sizes of $(2,4)$ were explored and limited by GPU memory. Input token lengths explored were $(800,1024)$. All hyperparameter searching was done on the dev set.

The BERT-based models converge between 0-10 epochs as well and take from around 1-3 hours to converge. Weighted average parameter $\alpha$ from $(0,0.1,0.2,0.25,0.5,0.6,0.7,0.75,0.8,0.9,1)$. Combinations of learning rate and batch size $(1e$-$4,2e$-$4)\times (256,512,1024)$ were exhaustively searched. Optimal learning rate, batch size pair  $=(2e$-$4,512)$.






\subsection{Non-author Human Expert Generated Slides}
Table \ref{tab:human-four} reports mean ROUGE F-scores (standard deviation in bracket) for non-author generated slides in comparison to our keyword model. Paper 960 was annotated by three human experts in order to measure human performance similarity. Although human experts outperform all systems by a large margin in terms of readability, informativeness, and consistency (see Figure \ref{fig:ratings}), it seems that our model is comparable to and sometimes surpasses human performance regarding finding different pieces of relevant information.


\subsection{Human Evaluation Survey System}
We designed and implemented a web-based survey system to support the human evaluation study, as presented in the Human Evaluate section in the main text. Figure~\ref{fig:survey_screenshot} shows a screenshot of the survey. The original slide deck was displayed at the top, along with a link to the original paper. This is to make sure that participants have everything they need to understand the slide content. In total, 14 participants said they referred to the original papers a few times.

In each round, the participant was given one or more original slides with the same title as reference and was asked to evaluate the corresponding slides generated by the three models, as well as those from non-authors when available. 
The model names were hidden from the participants and the order of the methods were also randomized across rounds to ensure each round is evaluated independently without bias.

During the evaluation process, 
the participant could use the three buttons below the slide image to flip through the deck or go back to the slide that was under inspection for the current round. That slide was also shown at the beginning of each round. Occasionally, multiple original slides had the same title and they all need to be inspected. The participant was notified about this situation via a flashing highlighted message on top of the original slide image, as shown in the screenshot.

The bottom section contains the two tasks that the participants needed to complete. The first one contains the rating tasks and the second one the ranking task. The participant can only go to the next round after all tasks were completed. Participants were told that the model numbers can change from round to round, and that mentions of tables or figures should be ignored.

\subsection{Slides Generated From Different Systems}
Table \ref{tab:slidesExample} shows a few examples of slide content generated from different systems under the same title. Colored text snippets indicate that the 
generated contents are similar to the original slide contents.

\begin{table*}[t]

\begin{small}
\begin{tabular}{|p{2.5cm}|p{12cm}|}
\hline
 \multicolumn{2}{|c|} {Slide Title: \emph{\textbf{\color{blue}{Motivations}}}}\\ \hline
\textbf{Original Content} & [s1] \textcolor{orange}{Insufficient or even unavailable training data} of emerging classes is a big challenge in real-world text classification. [s2] \textcolor{red}{Zero-shot text classification recognising text documents of classes that have never been seen in the learning stage.} [s3] In this paper, we propose a two-phase framework together with data augmentation and feature augmentation to solve this problem.\\ \hline
\textbf{BARTKeyword} & [s1] \textcolor{red}{Zero-shot learning (ZSL) for text classification aims to classify documents of classes which are absent from the learning stage.} [s2] The number of topics on social media is growing rapidly, and the classification models are required to recognise the text of the new topics using only general information (e.g., descriptions of the topics) since \textcolor{orange}{labelled training instances are unfeasible.}\\ \hline
\textbf{BARTSumm} & [s1] \textcolor{red}{Zero-shot learning (ZSL) for text classification aims to classify documents of classes which are absent from the learning stage.} [s2] Humans are able to learn new concepts by transferring knowledge from known to unknown domains based on high-level descriptions and semantic representations (Thrun and Pratt, 1998). [s3] ZSL is expected to exploit supportive semantic knowledge (e.g., class descriptions, relations among classes, and external domain knowledge) to generally infer the features of unseen classes using patterns learned from seen classes.\\ \hline
\textbf{BertSummExt} & [s1] The most widely used one is semantic attributes of classes such as visual concepts (e.g., colours, shapes) and semantic properties (e.g., behaviours, functions) (Lampert et al., 2009; Zhao et al., 2018). [s2] Hence, when the names of unseen classes are known in the inference stage, we try to introduce them to the classifiers in Phase 1 via augmented data so they can learn to reject the instances likely from unseen classes. [s3] So far, three main types of semantic knowledge have been employed in general zero-shot scenarios.\\ \hline
 \multicolumn{2}{|c|} {Slide Title: \emph{\textbf{\color{blue}{Our Proposed Framework Overview}}}}\\ \hline
\textbf{Original Content} & [s1] We integrate four kinds of semantic knowledge into our framework. [s2] \textcolor{red}{Data augmentation technique helps the classifiers be aware of the existence of unseen classes without accessing their real data.} [s3] \textcolor{orange}{Feature augmentation} provides additional information which relates the document and the unseen classes to generalise the zero-shot reasoning.\\ \hline
\textbf{BARTKeyword} & [s1] We propose a novel deep learning based Two phase framework to tackle the zero-shot text classification problem. [s2] The first phase predicts whether an input document comes from seen or unseen classes. [s3] \textcolor{red}{We also apply a data augmentation technique in this phase to help the classifiers be aware of the existence of unseen classes without accessing their real data.} [s4] Then the second phase, fine-grained classification, finally specifies the class of the input document.\\ \hline
\textbf{BARTSumm} & [s1] We propose \textcolor{orange}{a novel feature augmentation technique} called topic translation to detect documents from unseen classes by topic-translation from the documents of seen classes in the training dataset. [s2] In the document level, we follow Algorithm 1 to translate a document of class c into the topic of another class c. [s3] To explain, we translate all nouns, verbs, adjectives, and adverbs in the given document to the target class, word-by-word, using the word-level analogy.\\ \hline
\textbf{BertSummExt} & [s1] All notations are defined in section 2.1-2.2. [s2] For example, solving the analogy \"company:firm :: village:?\" [s3] The fine-grained phase finally decides the class $y_{i}$.\\ \hline
 \multicolumn{2}{|c|} {Slide Title: \emph{\textbf{\color{blue}{Experiments data sets}}}}\\ \hline
\textbf{Original Content} & [s1] \textcolor{red}{Spanish and Dutch: CoNLL 2002} [s2] \textcolor{orange}{Russian: LDC2016E95 (Russian Representative Language Pack)} [s3] \textcolor{magenta}{Chechen: TAC KBP 2017 10-Language EDL Pilot Evaluation Source Corpus} [s4] Part-of-speech Tagging: CoNLL 2017 (Universal Dependencies)\\ \hline
\textbf{BARTKeyword} & [s1] We use the following data sets: \textcolor{red}{Dutch (NLD) and Spanish (ESP) data from the CoNLL 2002 shared task} (Tjong Kim Sang, 2002). [s2] English (ENG) and English (ENG), \textcolor{orange}{Russian (RUS) Data from LDC2016E95} (Russian Representative Language Pack), and \textcolor{magenta}{Chechen (CHE) Data from TAC KBP 2017 10-Language EDL Pilot Evaluation Source Corpus} 4.\\ \hline
\textbf{BARTSumm} & [s1] LSTM-CRFs-based models are well-suited for multi-lingual multi-task learning for three reasons: (1) They learn features from word and character embeddings and therefore require little feature engineering; (2) As the input and output of each layer in a neural network are abstracted as vectors, it is fairly straightforward to share components between neural models; (3) Character Embeddings can serve as a bridge to transfer morphological and semantic information between languages with identical or similar scripts, without requiring crosslingual dictionaries or parallel sentences.\\ \hline
\textbf{BertSummExt} & [s1] Experiments Data Sets For Name Tagging, we use the following data sets: \textcolor{red}{Dutch (NLD) and Spanish (ESP) data from the CoNLL 2002 shared task (Tjong Kim Sang, 2002)}, English (ENG) data from the CoNLL 2003 shared task (Tjong Kim Sang and De Meulder, 2003), \textcolor{orange}{Russian (RUS) data from LDC2016E95} (Russian Representative Language Pack), and \textcolor{magenta}{Chechen (CHE) data from TAC KBP 2017 10-Language EDL Pilot Evaluation Source Corpus} 4. [s2] In this data set, each token is annotated with two POS tags, UPOS (universal POS tag) and XPOS (language-specific POS tag). [s3] English, Spanish, and Dutch embeddings are trained on corresponding Wikipedia articles (2017-12-20 dumps).\\ \hline

\end{tabular}
\end{small}
\caption{Slides generated from different systems.}
\label{tab:slidesExample}
\end{table*}

%% file: d2s_main_final.bbl
\begin{thebibliography}{44}
\expandafter\ifx\csname natexlab\endcsname\relax\def\natexlab#1{#1}\fi

\bibitem[{GRO(2008--2020)}]{GROBID}
 2008--2020.
\newblock \href
  {http://arxiv.org/abs/1:dir:dab86b296e3c3216e2241968f0d63b68e8209d3c}
  {Grobid}.
\newblock \url{https://github.com/kermitt2/grobid}.

\bibitem[{Al~Masum et~al.(2005)Al~Masum, Ishizuka, and Islam}]{al2005auto}
Shaikh~Mostafa Al~Masum, Mitsuru Ishizuka, and Md~Tawhidul Islam. 2005.
\newblock 'auto-presentation': a multi-agent system for building automatic
  multi-modal presentation of a topic from world wide web information.
\newblock In \emph{IEEE/WIC/ACM International Conference on Intelligent Agent
  Technology}, pages 246--249. IEEE.

\bibitem[{Bartsch and Cobern(2003)}]{bartsch2003effectiveness}
Robert~A Bartsch and Kristi~M Cobern. 2003.
\newblock Effectiveness of powerpoint presentations in lectures.
\newblock \emph{Computers \& education}, 41(1):77--86.

\bibitem[{Baumel et~al.(2016)Baumel, Cohen, and
  Elhadad}]{10.5555/3016100.3016261}
Tal Baumel, Raphael Cohen, and Michael Elhadad. 2016.
\newblock Topic concentration in query focused summarization datasets.
\newblock In \emph{Proceedings of the Thirtieth AAAI Conference on Artificial
  Intelligence}, AAAI'16, page 2573–2579. AAAI Press.

\bibitem[{Bhandare et~al.(2016)Bhandare, Awati, and
  Kharade}]{bhandare2016automatic}
Anuja~A Bhandare, Chetan~J Awati, and Sonam~S Kharade. 2016.
\newblock Automatic era: Presentation slides from academic paper.
\newblock In \emph{2016 International Conference on Automatic Control and
  Dynamic Optimization Techniques (ICACDOT)}, pages 809--814. IEEE.

\bibitem[{Cachola et~al.(2020)Cachola, Lo, Cohan, and
  Weld}]{cachola-etal-2020-tldr}
Isabel Cachola, Kyle Lo, Arman Cohan, and Daniel Weld. 2020.
\newblock \href {https://www.aclweb.org/anthology/2020.findings-emnlp.428}
  {{TLDR}: Extreme summarization of scientific documents}.
\newblock In \emph{Findings of the Association for Computational Linguistics:
  EMNLP 2020}, pages 4766--4777, Online. Association for Computational
  Linguistics.

\bibitem[{Clark and Divvala(2016)}]{clark2016pdffigures}
Christopher Clark and Santosh Divvala. 2016.
\newblock Pdffigures 2.0: Mining figures from research papers.
\newblock In \emph{2016 IEEE/ACM Joint Conference on Digital Libraries (JCDL)},
  pages 143--152. IEEE.

\bibitem[{Dang(2005)}]{Dang2005OverviewOD}
H.~Dang. 2005.
\newblock Overview of duc 2005.

\bibitem[{Erera et~al.(2019)Erera, Shmueli-Scheuer, Feigenblat, Peled~Nakash,
  Boni, Roitman, Cohen, Weiner, Mass, Rivlin, Lev, Jerbi, Herzig, Hou, Jochim,
  Gleize, Bonin, Bonin, and Konopnicki}]{erera-etal-2019-summarization}
Shai Erera, Michal Shmueli-Scheuer, Guy Feigenblat, Ora Peled~Nakash, Odellia
  Boni, Haggai Roitman, Doron Cohen, Bar Weiner, Yosi Mass, Or~Rivlin, Guy Lev,
  Achiya Jerbi, Jonathan Herzig, Yufang Hou, Charles Jochim, Martin Gleize,
  Francesca Bonin, Francesca Bonin, and David Konopnicki. 2019.
\newblock \href {https://doi.org/10.18653/v1/D19-3036} {A summarization system
  for scientific documents}.
\newblock In \emph{Proceedings of the 2019 Conference on Empirical Methods in
  Natural Language Processing and the 9th International Joint Conference on
  Natural Language Processing (EMNLP-IJCNLP): System Demonstrations}, pages
  211--216, Hong Kong, China. Association for Computational Linguistics.

\bibitem[{Fan et~al.(2019)Fan, Jernite, Perez, Grangier, Weston, and
  Auli}]{fan-etal-2019-eli5}
Angela Fan, Yacine Jernite, Ethan Perez, David Grangier, Jason Weston, and
  Michael Auli. 2019.
\newblock \href {https://doi.org/10.18653/v1/P19-1346} {{ELI}5: Long form
  question answering}.
\newblock In \emph{Proceedings of the 57th Annual Meeting of the Association
  for Computational Linguistics}, pages 3558--3567, Florence, Italy.
  Association for Computational Linguistics.

\bibitem[{Feigenblat et~al.(2017)Feigenblat, Roitman, Boni, and
  Konopnicki}]{10.1145/3077136.3080690}
Guy Feigenblat, Haggai Roitman, Odellia Boni, and David Konopnicki. 2017.
\newblock \href {https://doi.org/10.1145/3077136.3080690} {Unsupervised
  query-focused multi-document summarization using the cross entropy method}.
\newblock In \emph{Proceedings of the 40th International ACM SIGIR Conference
  on Research and Development in Information Retrieval}, SIGIR '17, page
  961–964, New York, NY, USA. Association for Computing Machinery.

\bibitem[{Field(2009)}]{field2009discovering}
Andy Field. 2009.
\newblock \emph{Discovering statistics using SPSS:(and sex and drugs and
  rock'n'roll)}.
\newblock Sage.

\bibitem[{Franco et~al.(2016)Franco, Thirumoorthy, and
  Muneeswaran}]{franco2016automatic}
P~Regnath Franco, K~Thirumoorthy, and K~Muneeswaran. 2016.
\newblock Automatic creation of well-organized slides from documents.
\newblock In \emph{2016 International Conference on Wireless Communications,
  Signal Processing and Networking (WiSPNET)}, pages 1173--1177. IEEE.

\bibitem[{Gehrmann et~al.(2018)Gehrmann, Deng, and
  Rush}]{gehrmann-etal-2018-bottom}
Sebastian Gehrmann, Yuntian Deng, and Alexander Rush. 2018.
\newblock \href {https://doi.org/10.18653/v1/D18-1443} {Bottom-up abstractive
  summarization}.
\newblock In \emph{Proceedings of the 2018 Conference on Empirical Methods in
  Natural Language Processing}, pages 4098--4109, Brussels, Belgium.
  Association for Computational Linguistics.

\bibitem[{Grusky et~al.(2018)Grusky, Naaman, and
  Artzi}]{grusky-etal-2018-newsroom}
Max Grusky, Mor Naaman, and Yoav Artzi. 2018.
\newblock \href {https://doi.org/10.18653/v1/N18-1065} {{N}ewsroom: A dataset
  of 1.3 million summaries with diverse extractive strategies}.
\newblock In \emph{Proceedings of the 2018 Conference of the North {A}merican
  Chapter of the Association for Computational Linguistics: Human Language
  Technologies, Volume 1 (Long Papers)}, pages 708--719, New Orleans,
  Louisiana. Association for Computational Linguistics.

\bibitem[{Guu et~al.(2020)Guu, Lee, Tung, Pasupat, and Chang}]{Guu2020REALMRL}
Kelvin Guu, Kenton Lee, Zora Tung, Panupong Pasupat, and Ming-Wei Chang. 2020.
\newblock Realm: Retrieval-augmented language model pre-training.
\newblock \emph{ArXiv}, abs/2002.08909.

\bibitem[{Hermann et~al.(2015)Hermann, Kocisky, Grefenstette, Espeholt, Kay,
  Suleyman, and Blunsom}]{NIPS2015_afdec700}
Karl~Moritz Hermann, Tomas Kocisky, Edward Grefenstette, Lasse Espeholt, Will
  Kay, Mustafa Suleyman, and Phil Blunsom. 2015.
\newblock \href
  {https://proceedings.neurips.cc/paper/2015/file/afdec7005cc9f14302cd0474fd0f3c96-Paper.pdf}
  {Teaching machines to read and comprehend}.
\newblock In \emph{Advances in Neural Information Processing Systems},
  volume~28, pages 1693--1701. Curran Associates, Inc.

\bibitem[{Hou et~al.(2019)Hou, Jochim, Gleize, Bonin, and
  Ganguly}]{hou-etal-2019-identification}
Yufang Hou, Charles Jochim, Martin Gleize, Francesca Bonin, and Debasis
  Ganguly. 2019.
\newblock \href {https://doi.org/10.18653/v1/P19-1513} {Identification of
  tasks, datasets, evaluation metrics, and numeric scores for scientific
  leaderboards construction}.
\newblock In \emph{Proceedings of the 57th Annual Meeting of the Association
  for Computational Linguistics}, pages 5203--5213, Florence, Italy.
  Association for Computational Linguistics.

\bibitem[{Hu and Wan(2014)}]{hu2014ppsgen}
Yue Hu and Xiaojun Wan. 2014.
\newblock Ppsgen: Learning-based presentation slides generation for academic
  papers.
\newblock \emph{IEEE transactions on knowledge and data engineering},
  27(4):1085--1097.

\bibitem[{Johnson et~al.(2019)Johnson, Douze, and
  J{\'e}gou}]{johnson2019billion}
Jeff Johnson, Matthijs Douze, and Herv{\'e} J{\'e}gou. 2019.
\newblock Billion-scale similarity search with gpus.
\newblock \emph{IEEE Transactions on Big Data}.

\bibitem[{Kan(2007)}]{kan2007slideseer}
Min-Yen Kan. 2007.
\newblock Slideseer: A digital library of aligned document and presentation
  pairs.
\newblock In \emph{Proceedings of the 7th ACM/IEEE-CS joint conference on
  Digital libraries}, pages 81--90.

\bibitem[{Karpukhin et~al.(2020)Karpukhin, Oguz, Min, Lewis, Wu, Edunov, Chen,
  and Yih}]{karpukhin-etal-2020-dense}
Vladimir Karpukhin, Barlas Oguz, Sewon Min, Patrick Lewis, Ledell Wu, Sergey
  Edunov, Danqi Chen, and Wen-tau Yih. 2020.
\newblock \href {https://www.aclweb.org/anthology/2020.emnlp-main.550} {Dense
  passage retrieval for open-domain question answering}.
\newblock In \emph{Proceedings of the 2020 Conference on Empirical Methods in
  Natural Language Processing (EMNLP)}, pages 6769--6781, Online. Association
  for Computational Linguistics.

\bibitem[{Kryscinski et~al.(2019)Kryscinski, Keskar, McCann, Xiong, and
  Socher}]{kryscinski-etal-2019-neural}
Wojciech Kryscinski, Nitish~Shirish Keskar, Bryan McCann, Caiming Xiong, and
  Richard Socher. 2019.
\newblock \href {https://doi.org/10.18653/v1/D19-1051} {Neural text
  summarization: A critical evaluation}.
\newblock In \emph{Proceedings of the 2019 Conference on Empirical Methods in
  Natural Language Processing and the 9th International Joint Conference on
  Natural Language Processing (EMNLP-IJCNLP)}, pages 540--551, Hong Kong,
  China. Association for Computational Linguistics.

\bibitem[{Kwiatkowski et~al.(2019)Kwiatkowski, Palomaki, Redfield, Collins,
  Parikh, Alberti, Epstein, Polosukhin, Devlin, Lee, Toutanova, Jones, Kelcey,
  Chang, Dai, Uszkoreit, Le, and Petrov}]{kwiatkowski-etal-2019-natural}
Tom Kwiatkowski, Jennimaria Palomaki, Olivia Redfield, Michael Collins, Ankur
  Parikh, Chris Alberti, Danielle Epstein, Illia Polosukhin, Jacob Devlin,
  Kenton Lee, Kristina Toutanova, Llion Jones, Matthew Kelcey, Ming-Wei Chang,
  Andrew~M. Dai, Jakob Uszkoreit, Quoc Le, and Slav Petrov. 2019.
\newblock \href {https://doi.org/10.1162/tacl_a_00276} {Natural questions: A
  benchmark for question answering research}.
\newblock \emph{Transactions of the Association for Computational Linguistics},
  7:453--466.

\bibitem[{Lee et~al.(2019)Lee, Chang, and Toutanova}]{lee2019latent}
Kenton Lee, Ming-Wei Chang, and Kristina Toutanova. 2019.
\newblock Latent retrieval for weakly supervised open domain question
  answering.
\newblock In \emph{Proceedings of the 57th Annual Meeting of the Association
  for Computational Linguistics}, pages 6086--6096.

\bibitem[{Lewis et~al.(2020)Lewis, Liu, Goyal, Ghazvininejad, Mohamed, Levy,
  Stoyanov, and Zettlemoyer}]{lewis-etal-2020-bart}
Mike Lewis, Yinhan Liu, Naman Goyal, Marjan Ghazvininejad, Abdelrahman Mohamed,
  Omer Levy, Veselin Stoyanov, and Luke Zettlemoyer. 2020.
\newblock \href {https://doi.org/10.18653/v1/2020.acl-main.703} {{BART}:
  Denoising sequence-to-sequence pre-training for natural language generation,
  translation, and comprehension}.
\newblock In \emph{Proceedings of the 58th Annual Meeting of the Association
  for Computational Linguistics}, pages 7871--7880, Online. Association for
  Computational Linguistics.

\bibitem[{Lin(2004)}]{lin-2004-rouge}
Chin-Yew Lin. 2004.
\newblock \href {https://www.aclweb.org/anthology/W04-1013} {{ROUGE}: A package
  for automatic evaluation of summaries}.
\newblock In \emph{Text Summarization Branches Out}, pages 74--81, Barcelona,
  Spain. Association for Computational Linguistics.

\bibitem[{Liu and Lapata(2019)}]{liu2019text}
Yang Liu and Mirella Lapata. 2019.
\newblock \href {https://doi.org/10.18653/v1/D19-1387} {Text summarization with
  pretrained encoders}.
\newblock In \emph{Proceedings of the 2019 Conference on Empirical Methods in
  Natural Language Processing and the 9th International Joint Conference on
  Natural Language Processing (EMNLP-IJCNLP)}, pages 3730--3740, Hong Kong,
  China. Association for Computational Linguistics.

\bibitem[{Narayan et~al.(2018)Narayan, Cohen, and
  Lapata}]{narayan-etal-2018-dont}
Shashi Narayan, Shay~B. Cohen, and Mirella Lapata. 2018.
\newblock \href {https://doi.org/10.18653/v1/D18-1206} {Don{'}t give me the
  details, just the summary! topic-aware convolutional neural networks for
  extreme summarization}.
\newblock In \emph{Proceedings of the 2018 Conference on Empirical Methods in
  Natural Language Processing}, pages 1797--1807, Brussels, Belgium.
  Association for Computational Linguistics.

\bibitem[{Piorkowski et~al.(2021)Piorkowski, Park, Wang, Wang, Muller, and
  Portnoy}]{piorkowski2021ai}
David Piorkowski, Soya Park, April~Yi Wang, Dakuo Wang, Michael Muller, and
  Felix Portnoy. 2021.
\newblock How ai developers overcome communication challenges in a
  multidisciplinary team: A case study.
\newblock \emph{arXiv preprint arXiv:2101.06098}.

\bibitem[{Prasad et~al.(2009)Prasad, Mathivanan, Greetha, and
  Jayaprakasam}]{prasad2009document}
K~Gokul Prasad, Harish Mathivanan, TV~Greetha, and M~Jayaprakasam. 2009.
\newblock Document summarization and information extraction for generation of
  presentation slides.
\newblock In \emph{2009 International Conference on Advances in Recent
  Technologies in Communication and Computing}, pages 126--128. IEEE.

\bibitem[{Ribeiro et~al.(2020)Ribeiro, Wu, Guestrin, and
  Singh}]{ribeiro2020beyond}
Marco~Tulio Ribeiro, Tongshuang Wu, Carlos Guestrin, and Sameer Singh. 2020.
\newblock Beyond accuracy: Behavioral testing of nlp models with checklist.
\newblock \emph{arXiv preprint arXiv:2005.04118}.

\bibitem[{See et~al.(2017)See, Liu, and Manning}]{see-etal-2017-get}
Abigail See, Peter~J. Liu, and Christopher~D. Manning. 2017.
\newblock \href {https://doi.org/10.18653/v1/P17-1099} {Get to the point:
  Summarization with pointer-generator networks}.
\newblock In \emph{Proceedings of the 55th Annual Meeting of the Association
  for Computational Linguistics (Volume 1: Long Papers)}, pages 1073--1083,
  Vancouver, Canada. Association for Computational Linguistics.

\bibitem[{Sefid et~al.(2019)Sefid, Wu, Mitra, and Giles}]{sefid2019automatic}
Athar Sefid, Jian Wu, Prasenjit Mitra, and C~Lee Giles. 2019.
\newblock Automatic slide generation for scientific papers.
\newblock In \emph{SciKnow@ K-CAP}, pages 11--16.

\bibitem[{Shibata and Kurohashi(2005)}]{shibata2005automatic}
Tomohide Shibata and Sadao Kurohashi. 2005.
\newblock Automatic slide generation based on discourse structure analysis.
\newblock In \emph{International Conference on Natural Language Processing},
  pages 754--766. Springer.

\bibitem[{Syamili and Abraham(2017)}]{syamili2017presentation}
S~Syamili and Anish Abraham. 2017.
\newblock Presentation slides generation from scientific papers using support
  vector regression.
\newblock In \emph{2017 International Conference on Inventive Communication and
  Computational Technologies (ICICCT)}, pages 286--291. IEEE.

\bibitem[{Turc et~al.(2019)Turc, Chang, Lee, and
  Toutanova}]{Turc2019WellReadSL}
Iulia Turc, Ming-Wei Chang, Kenton Lee, and Kristina Toutanova. 2019.
\newblock Well-read students learn better: On the importance of pre-training
  compact models.
\newblock \emph{arXiv: Computation and Language}.

\bibitem[{Wang(2016)}]{wang2016people}
Dakuo Wang. 2016.
\newblock How people write together now: Exploring and supporting today's
  computer-supported collaborative writing.
\newblock In \emph{Proceedings of the 19th ACM Conference on Computer Supported
  Cooperative Work and Social Computing Companion}, pages 175--179.

\bibitem[{Wang et~al.(2021)Wang, Wang, Tian, Peng, Fan, Zhang, Ma, Yu, Ma, and
  Wang}]{wang2021cass}
Liuping Wang, Dakuo Wang, Feng Tian, Zhenhui Peng, Xiangmin Fan, Zhan Zhang,
  Shuai Ma, Mo~Yu, Xiaojuan Ma, and Hongan Wang. 2021.
\newblock Cass: Towards building a social-support chatbot for online health
  community.
\newblock \emph{CSCW'21}.

\bibitem[{Wang et~al.(2017)Wang, Wan, and Du}]{wang2017phrase}
Sida Wang, Xiaojun Wan, and Shikang Du. 2017.
\newblock Phrase-based presentation slides generation for academic papers.
\newblock In \emph{Proceedings of the Thirty-First AAAI Conference on
  Artificial Intelligence}, pages 196--202.

\bibitem[{Wang and Sumiya(2013)}]{wang2013method}
Yuanyuan Wang and Kazutoshi Sumiya. 2013.
\newblock A method for generating presentation slides based on expression
  styles using document structure.
\newblock \emph{International Journal of Knowledge and Web Intelligence},
  4(1):93--112.

\bibitem[{Winters and Mathewson(2019)}]{winters2019automatically}
Thomas Winters and Kory~W Mathewson. 2019.
\newblock Automatically generating engaging presentation slide decks.
\newblock In \emph{International Conference on Computational Intelligence in
  Music, Sound, Art and Design (Part of EvoStar)}, pages 127--141. Springer.

\bibitem[{Wolf et~al.(2020)Wolf, Chaumond, Debut, Sanh, Delangue, Moi, Cistac,
  Funtowicz, Davison, Shleifer et~al.}]{wolf2020transformers}
Thomas Wolf, Julien Chaumond, Lysandre Debut, Victor Sanh, Clement Delangue,
  Anthony Moi, Pierric Cistac, Morgan Funtowicz, Joe Davison, Sam Shleifer,
  et~al. 2020.
\newblock Transformers: State-of-the-art natural language processing.
\newblock In \emph{Proceedings of the 2020 Conference on Empirical Methods in
  Natural Language Processing: System Demonstrations}, pages 38--45.

\bibitem[{Xu and Lapata(2020)}]{xu-lapata-2020-coarse}
Yumo Xu and Mirella Lapata. 2020.
\newblock \href {https://www.aclweb.org/anthology/2020.emnlp-main.296}
  {Coarse-to-fine query focused multi-document summarization}.
\newblock In \emph{Proceedings of the 2020 Conference on Empirical Methods in
  Natural Language Processing (EMNLP)}, pages 3632--3645, Online. Association
  for Computational Linguistics.

\end{thebibliography}
